\documentclass{llncs}
\usepackage{llncsdoc}
\usepackage[pdftex]{graphicx}

\usepackage{color}
\usepackage{tikz}
\usepackage{pgfplots}
\usepackage{pgf-umlsd}
\usepackage{ifthen}
\usepackage{cite}

\usepackage{makecell}

\begin{document}

\title{Multi Modal Convolutional Neural Networks for Brain Tumor Segmentation }


\author{Mehmet Ayg\"{u}n,    Yusuf H\"{u}seyin \c{S}ahin,  G\"{o}zde \"{U}nal  \\ \{aygunme, sahinyu, unalgo\}@itu.edu.tr}

\institute{ Istanbul Technical University \\ Istanbul, Turkey }

\maketitle
\begin{abstract}
In this work, we propose a multi-modal Convolutional Neural Network (CNN) approach for brain tumor segmentation. We investigate how to combine different modalities efficiently in the CNN framework. We adapt various fusion methods, which are previously employed on video recognition problem, to the brain tumor segmentation problem, and we investigate their efficiency in terms of memory and performance. Our experiments, which are performed on BRATS dataset, lead us to the conclusion that learning separate  representations for each modality and combining them for brain tumor segmentation could increase the performance of CNN systems.
\end{abstract}
\section{Introduction}
The magnetic resonance imaging (MRI) technique uses magnetic field and radio waves to capture detailed images of the brain. Finding tumor areas automatically in the brain could help in improved diagnostics, treatment planning, and growth rate prediction. Recent developments in learning-based vision algorithms led to promising results in image-based tumor analysis. For instance, latest methods achieve around 90\% accuracy in the challenging multi-modal brain tumor segmentation benchmark named BRATS \cite{menze2015multimodal}. Segmenting tumor areas in the brain is a very challenging task due to the huge variety in shape and location of tumor tissue in the brain.\par On the other hand, CNNs have become default method for the segmentation task in medical images based on their recent success on various computer vision tasks \cite{long2015fully,he2016deep,redmon2016you}. However, due to computational complexity and memory fingerprint of the CNNs, they need to be adapted for the medical image segmentation problem. We provide a detailed overview of recent works for medical image segmentation problem in the next section. \par
\begin{figure}
\begin{center}
        \includegraphics[width=\linewidth]{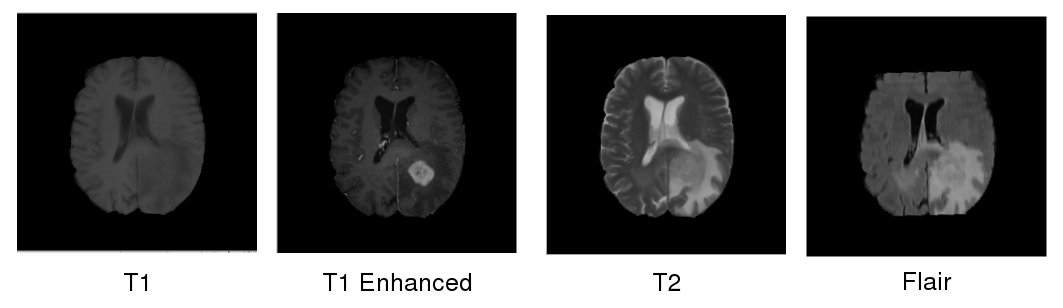}
        \caption{Different modalities for a sample patient data from BRATS dataset \cite{menze2015multimodal}. Each modality provides different information that complement each other. }
        \label{fig:multimodal}
    \end{center}         
\end{figure}

 Whereas earlier methods mainly aim to capture 3 dimensional structure of the anatomy using a single modality, the MRI devices provide different image contrasts with different gray scale intensities. For instance, in the BRATS dataset\cite{menze2015multimodal}, for each patient four different modality images such as T1-weighted MRI, T1-weighted enhanced MRI, T2-weighted MRI and FLAIR are acquired. As can be seen in Figure \ref{fig:multimodal} all of these modalities provide different information about the brain anatomy of the patient. However, in the recent works, these modalities are generally just concatenated, and fed to CNNs as an input. Thus, the CNNs learn a single representation for all modalities. In this work, we adapt a number of fusion methods that were previously applied on video action recognition \cite{feichtenhofer2016convolutional} problem for learning separate representations for each modality, and we combine these modalities efficiently in the CNN framework for an improved tumor segmentation. Our experiments show that with appropriate fusion method and fusion point in the CNN architecture, the error rate of the CNNs could be reduced by 30\%. 

\section{Related Work}

The CNNs are adapted for medical image segmentation problem because of their recent success on various computer vision task. But, two dimensional CNNs cannot br applied directly to medical domain since medical images are generally three dimensional. Various recent works aimed at adapting CNNs for medical image segmentation problem. The most basic approach was based on the idea of taking 2D slices from scan of patient, feeding to network and predicting 2D output maps. However, as expected in those 2D approaches, 3D dimensional structure of the anatomy cannot be learned. For capturing the 3 dimensional structure, different methods are suggested such as U-Net like architectures \cite{ronneberger2015u,cciccek20163d}, 3D CNNs \cite{kamnitsas2017efficient} and Recurrent Networks \cite{chen2016combining,poudel2016recurrent}.

\par
\begin{figure}
\begin{center}
        \includegraphics[width=0.7\linewidth]{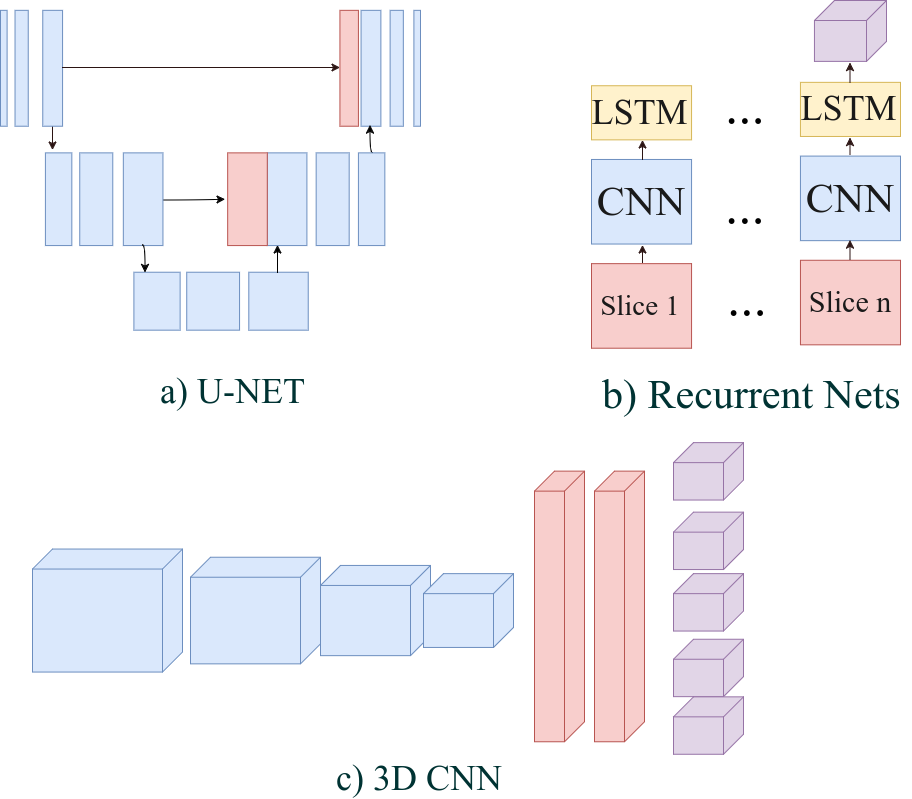}
        \caption{Different CNN-based approaches for volumetric medical image segmentation. The U-Net networks \cite{ronneberger2015u,cciccek20163d} (a) are similar to FCN \cite{long2015fully} networks and the 3D Patch based CNNs \cite{kamnitsas2017efficient} (c) are very much alike regular feed forward networks. The Recurrent networks \cite{chen2016combining,poudel2016recurrent} (b) use idea of capturing temporal information with state connections.}
        \label{fig:rl}
    \end{center}         
\end{figure}

 The U-Net architectures \cite{ronneberger2015u} are like U shape networks. The shape of input and output is the same in the U-Net architecture, however differently from fully convolutional nets \cite{long2015fully}, they concatenate early layer outputs at later layers with the aim of bringing information that was lost in down-sampling operation. While the first version of U-Net used 2D scan slices, for using 3D scans, 3D U-Net \cite{cciccek20163d} was proposed. The second direction that researchers follow is using recurrent nets. In these approaches \cite{chen2016combining,poudel2016recurrent}, each slice is input to a CNN and the output of the CNN is input to an LSTM \cite{hochreiter1997long} unit which is a variant of recurrent nets. The output of the LSTM methods would be a 3D segmentation map of the scan in contrast to that of the 2D feedforward CNNs. In a last group of approaches, small 3D image patches are created from a volume and segmentation maps of those patches are generated \cite{kamnitsas2017efficient}. Since the patches are small, normal feed-forward CNNs with 3D convolutional layers can be used. Figure \ref{fig:rl} depicts visualizations of the architectures for the above-mentioned approaches. \par In this paper, as our aim is to explore ways of combining modalities, we adapt a 3D CNN that takes the 3D patch-based approach, since it is easy to modify and gives the best performance on recent segmentation tasks \cite{kamnitsas2017efficient}.

\section{Method}

\begin{figure}
\begin{center}
        \includegraphics[width=0.8\linewidth]{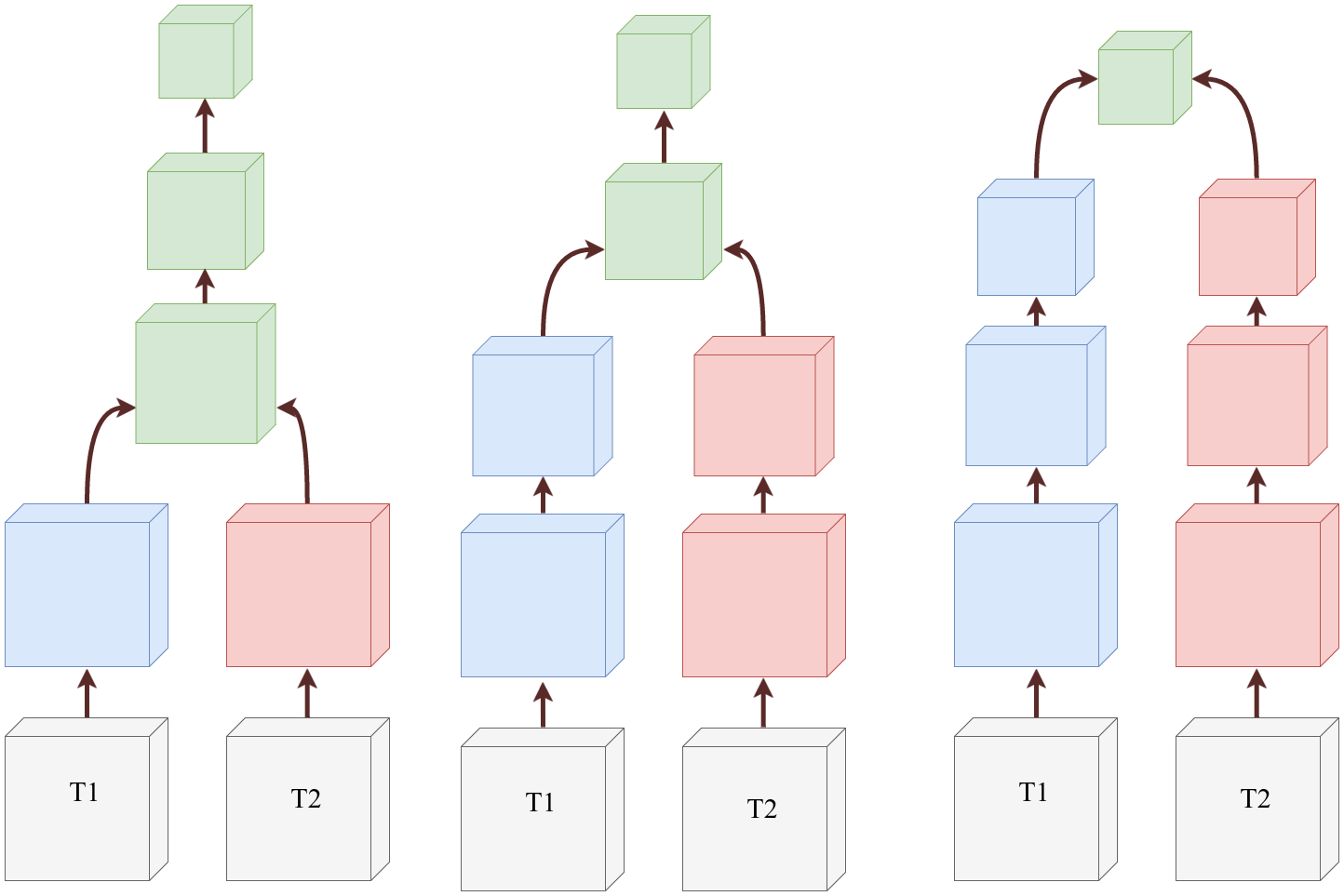}
        \caption{Different fusion points. Early, middle and late fusion point are shown from left to right, respectively. The exact definition of early, middle and late is subject to architecture that used in the problem. For our experiments we give the details in the experiment section. }
        \label{fig:fusion_point}
    \end{center}         
\end{figure}

In this section, the novel methods that we propose for combining modalities in CNNs are explained. Differently from previous works, we consider all modalities as separate inputs to different CNNs. However, the outputs of the CNNs need to combined for final prediction. For end to end training we concatenate these CNNs and produce a single output. Here, there are two problems that need to be solved : i) where to combine the CNNs and ii) how to combine these CNNs. For investigating those issues, we adapt a previous method that was proposed for action recognition problem \cite{feichtenhofer2016convolutional}. In the paper, the authors combine spatial and temporal CNNs with different fusion points and different fusion functions. We also employ a similar mechanism for brain tumor segmentation. Different from the earlier method, we employ four different CNNs for each modality in our method. Another difference is that as capturing 3 dimensional structure of the anatomy is important we use 3D CNNs for each modality. We employ the architecture that was described in the \cite{kamnitsas2017efficient} as a baseline model. The architecture takes an $25\times25\times25$ patch as a input and predicts a segmentation map of the $9\times9\times9$ area that is the center of the input patch. 
\par We define three different combining points: "early", "middle", and "late". Visualizations of these different fusion points can be seen in Figure \ref{fig:fusion_point}. In the "early" version, the CNNs are joined after early layers of the separate CNNs whereas in the "late" version, they are combined in the later layers. The specific definition of "early" or "late" depends on the utilized architecture. Furthermore, we define three different fusion functions for combining these CNNs. The first fusion function is maximum fusion, which is defined as:
\begin{equation}
Y(i,j,k) = max(X_{1}(i,j,k), X_{2}(i,j,k), .... X_{n}(i,j,k)).
\end{equation}
The input of the fusion function is $N\times D\times W\times H$,where $N$ is the number of input modalities, and the corresponding output size is $1\times D\times W\times H$. The above function picks the maximum value among all modalities at a fusion location. The second fusion function is summation fusion, which sums values from different modalities at the fusion point, and produces a single output value. This function is defined as,
\begin{equation}
Y(i,j,k) = \sum_{i=1}^{n} X_{i}(i,j,k)
\end{equation}
These first two fusion functions are parameter free. Our last proposed function is convolution fusion. This function first concatenates four modalities and applies a convolution operation for producing a single output. Again, for this fusion function, the shape of the input and the output would be  $N\times D\times W\times H$ and $1\times D\times W\times H$. We learn parameters of the convolution kernel in the training phase. To sum up, we employ three different fusion points and three different fusion functions. We experiment with each combination and report the performances on the brain tumor segmentation application in the next section.

\section{Experiments and Results}
\subsection{Experimental Setting}
We use BRATS 2015 \cite{menze2015multimodal} for our experiments. We perform our experiments on the training set, which is available for a quick experimentation. In the training set, there are 274 patients that are separated into two classes: low and high grade glioma patients (LGG and HGG). There are 54 low and 220 high grade patients. We split the whole training set into two subsets: (i) training subset is used for model training;  test subset is used for evaluation. We pick randomly 50 patients as a test set and we preserve  LGG-HGG ratio for train/test split. We use neither any data augmentation nor a pre-processing method beside normalizing each patient volume intensities to between range of 0 and 1. In addition to classification accuracy, we also report dice scores which is a more suitable measurement for tumor segmentation since the majority of the volumes are healthy.

\begin{figure}
	\noindent\resizebox{\textwidth}{!}{
	\begin{tikzpicture}
		\draw[use as bounding box, transparent] (-1.8,-1.8) rectangle (17.2, 3.2);


		\newcommand{\networkLayer}[6]{
			\def\a{#1} 
			\def\b{0.02}
			\def\c{#2} 
			\def\t{#3} 
			\ifthenelse {\equal{#6} {}} {\def\y{0}} {\def\y{#6}} 

			\draw[line width=0.25mm](\c+\t,0,0) -- (\c+\t,\a,0) -- (\t,\a,0);                                                      
			\draw[line width=0.25mm](\t,0,\a) -- (\c+\t,0,\a) node[midway,below] {#5} -- (\c+\t,\a,\a) -- (\t,\a,\a) -- (\t,0,\a); 
			\draw[line width=0.25mm](\c+\t,0,0) -- (\c+\t,0,\a);
			\draw[line width=0.25mm](\c+\t,\a,0) -- (\c+\t,\a,\a);
			\draw[line width=0.25mm](\t,\a,0) -- (\t,\a,\a);

			\filldraw[#4] (\t+\b,\b,\a) -- (\c+\t-\b,\b,\a) -- (\c+\t-\b,\a-\b,\a) -- (\t+\b,\a-\b,\a) -- (\t+\b,\b,\a); 
			\filldraw[#4] (\t+\b,\a,\a-\b) -- (\c+\t-\b,\a,\a-\b) -- (\c+\t-\b,\a,\b) -- (\t+\b,\a,\b);

			\ifthenelse {\equal{#4} {}}
			{} 
			{\filldraw[#4] (\c+\t,\b,\a-\b) -- (\c+\t,\b,\b) -- (\c+\t,\a-\b,\b) -- (\c+\t,\a-\b,\a-\b);} 
		}

		\networkLayer{2.5}{2.5}{-0.5}{color=green!50}{$4\times25\times25\times25$}

		\networkLayer{2.3}{2.3}{2.8}{color=blue!40}{conv1, 30}{$30\times21\times21\times21$}    
		\networkLayer{2.1}{2.1}{5.9}{color=blue!40}{conv2, 40}{$40\times17\times17\times17$}        
		\networkLayer{1.9}{1.9}{8.8}{color=blue!40}{conv3, 40}{$40\times13\times13\times13$}    
		\networkLayer{1.7}{1.7}{11.3}{color=blue!40}{conv4, 50}{$50\times9\times9\times9$}        


		\networkLayer{1.5}{1.5}{13.8}{color=red!40}{$5\times9\times9\times9$}          

	\end{tikzpicture}
	}
	\caption{Details of baseline model's architecture. All of the convolution layers are 3x3x3. The input (green) of the CNN is 4x25x25x25 which is concatenation of 25x25x25  patches that are extracted from different modalities. The output (red) is five 9x9x9 probability maps for each tumor class. Below the convolution filters (blue), name and depth of the layers are presented. Normally, there are eight convolution layers (i.e. conv1-1, conv1-2, conv2-1, conv2-2 etc.), but for visualization, only four of them are shown. }
	\label{fig:cnn}
\end{figure}
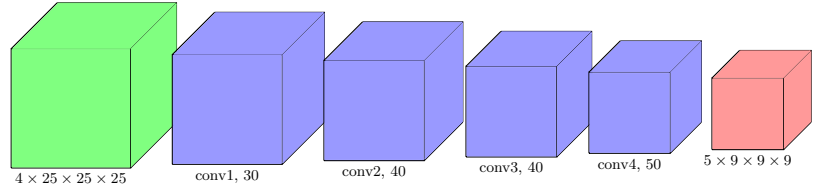

\subsection{Implementation Details}
We use Tensorflow \cite{tensorflow2015-whitepaper} for our implementations as Tensorflow already includes 3D convolution filters and allows an easy design of different architectures. We use Adam \cite{kingma2014adam} optimizer and Cross-Entropy loss for training our CNNs. Also for regularization, we apply L1 and L2 regularization on all weights and Dropout \cite{srivastava2014dropout} for fully connected and convolution layers. In the convolution layers, we use 2\% and in the FC 50\% probabilities for dropout layers. For all experiments, we train the models for 50 epochs, test the models on each epoch, and report the best validation accuracy.

\subsection{Results}
For comparing effect of multi-modal CNNs, first we perform experiments on a baseline model. We utilize the architecture introduced in \cite{kamnitsas2017efficient} as the baseline model. In the architecture, there are four convolution layers and two fully connected layers. Differently from their proposed method, we use only a single scale input with the shape of $25\times25\times25$, and do not employ any post-processing operation. The details of the architecture can be seen in Figure  \ref{fig:cnn}. In our settings, the baseline model gives 81.25\% dice and 98.20\% accuracy. For multi-modal training, we modify the baseline architecture. For the fusion point, we use end of first, second and fourth convolution layers for early, middle and late fusions, respectively. Also we experiment on the effect of different fusion functions. Results of all combinations can be seen in Table \ref{tbl:res}. 
\begin{table*}[t]
    
    \begin{center}
    \begin{tabular}{|c|r|r|} 
        \hline 
        \thead{Function} & \thead{ Dice } & \thead{Acc.}  \\ 
        \hline 
        sum &  78.04 & 98.33 \\ 
        \hline 
        max & 83.59 & 98.53 \\ 
        \hline 
        conv & 82.44 & 98.12 \\ 
        \hline 
    \end{tabular}
    \quad
    \begin{tabular}{|c|r|r|} 
        \hline 
        \thead{Function} & \thead{ Dice } & \thead{Acc.}  \\ 
        \hline 
        sum &  85.46 & 98.38 \\ 
        \hline 
        max & 81.99 & 98.44 \\ 
        \hline 
        conv & 78.25 & 98.13 \\ 
        \hline 
    \end{tabular}
    \quad
     \begin{tabular}{|c|r|r|} 
        \hline 
        \thead{Function} & \thead{ Dice } & \thead{Acc.}  \\ 
        \hline 
        sum &  85.00 & 98.40 \\ 
        \hline 
        max & 82.61 & 98.22 \\ 
        \hline 
        conv & 86.97 & 98.34 \\ 
        \hline 
    \end{tabular}

    \caption{Dice and Accuracy results for early fusion with different fusion functions. The tables are for early, middle and late fusion from left to right. The dice score is computed for whole tumor (all tumor classes).}
    \label{tbl:res}
    \end{center}  
\end{table*}
\par In terms of memory usage, the later fusion strategies have a memory handicap since the models fuse in later, they have more parameters. We also compare their accuracy gain/extra memory usage. We assume that accuracy / number of parameters ratio for baseline architecture as 1 and calculate the ratio for the other fusion strategies. The ratios can be seen in Table \ref{tbl:memory}. 

\begin{table}
    
    \begin{center}
    \begin{tabular}{|c|c|c|c|} 
        \hline 
        \thead{Function} & \thead{ Early } & \thead{ Middle}  & \thead{Late}  \\ 
        \hline 
        sum &  0.6209 & 0.3760 & 0.2636\\ 
        \hline 
        max & 0.6650 & 0.3607 & 0.2561\\ 
        \hline 
        conv & 0.5456 & 0.2872 & 0.2213 \\ 
        \hline 
    \end{tabular}
    \caption{Memory accuracy gain ratios for different fusion methods. We assumed accuracy/\#parameters of baseline method as 1 and calculate for other methods.}
    \label{tbl:memory}
    \end{center}  
\end{table}

\section{Discussion and Conclusion}
When we analyze the results in Table \ref{tbl:res}, except for early fusion with summation and middle fusion with convolution, all combinations outperform the baseline model. Moreover, for different fusion points, different fusion functions perform best, however in overall, late fusion method generally performs better. Late fusion with convolution gives the best result with 86.97 \% dice score, which is 5\% better than the baseline architecture. These results show that learning separate representations for each modality, which can be low level or high level, can increase the segmentation accuracy. Moreover, the later fusion provides better performance than the middle and early because while the early representations of the modalities can be similar, their representations at later layers learned with the CNNs reveal distinctive characteristics.
\par Although later fusion gives better performance, the memory efficiency of it is worse than early and middle fusion. This is because the fusion is executed in the later layers and separate convolutional layers are employed for each modality until the fusion is performed. Furthermore, the convolution fusion uses more memory as it requires convolution filters to perform convolution fusion. However the best accuracy is achieved with convolution fusion at later layers.
\par To summarize, in this work, we adapt various fusion methods for training multi-modal convolutional neural networks for brain tumor segmentation. In our experiments, learning different feature representations for each modality, increases the accuracy of the CNNs on challenging multi-modal brain tumor segmentation benchmark \cite{menze2015multimodal}. Furthermore, with a suitable fusion method, the error rate in dice overlap can be reduced by 30\%. 

\bibliographystyle{splncs03}
\bibliography{egbib} 

\begin{thebibliography}{10}
\providecommand{\url}[1]{\texttt{#1}}
\providecommand{\urlprefix}{URL }

\bibitem{tensorflow2015-whitepaper}
Abadi, M., Agarwal, A., Barham, P., Brevdo, E., Chen, Z., Citro, C., Corrado,
  G.S., Davis, A., Dean, J., Devin, M., Ghemawat, S., Goodfellow, I., Harp, A.,
  Irving, G., Isard, M., Jia, Y., Jozefowicz, R., Kaiser, L., Kudlur, M.,
  Levenberg, J., Man\'{e}, D., Monga, R., Moore, S., Murray, D., Olah, C.,
  Schuster, M., Shlens, J., Steiner, B., Sutskever, I., Talwar, K., Tucker, P.,
  Vanhoucke, V., Vasudevan, V., Vi\'{e}gas, F., Vinyals, O., Warden, P.,
  Wattenberg, M., Wicke, M., Yu, Y., Zheng, X.: {TensorFlow}: Large-scale
  machine learning on heterogeneous systems (2015),
  \url{http://tensorflow.org/}, software available from tensorflow.org

\bibitem{chen2016combining}
Chen, J., Yang, L., Zhang, Y., Alber, M., Chen, D.Z.: Combining fully
  convolutional and recurrent neural networks for 3d biomedical image
  segmentation. In: Advances in Neural Information Processing Systems. pp.
  3036--3044 (2016)

\bibitem{cciccek20163d}
{\c{C}}i{\c{c}}ek, {\"O}., Abdulkadir, A., Lienkamp, S.S., Brox, T.,
  Ronneberger, O.: 3d u-net: learning dense volumetric segmentation from sparse
  annotation. In: International Conference on Medical Image Computing and
  Computer-Assisted Intervention. pp. 424--432. Springer (2016)

\bibitem{feichtenhofer2016convolutional}
Feichtenhofer, C., Pinz, A., Zisserman, A.: Convolutional two-stream network
  fusion for video action recognition. In: Proceedings of the IEEE Conference
  on Computer Vision and Pattern Recognition. pp. 1933--1941 (2016)

\bibitem{he2016deep}
He, K., Zhang, X., Ren, S., Sun, J.: Deep residual learning for image
  recognition. In: Proceedings of the IEEE Conference on Computer Vision and
  Pattern Recognition. pp. 770--778 (2016)

\bibitem{hochreiter1997long}
Hochreiter, S., Schmidhuber, J.: Long short-term memory. Neural computation
  9(8),  1735--1780 (1997)

\bibitem{kamnitsas2017efficient}
Kamnitsas, K., Ledig, C., Newcombe, V.F., Simpson, J.P., Kane, A.D., Menon,
  D.K., Rueckert, D., Glocker, B.: Efficient multi-scale 3d cnn with fully
  connected crf for accurate brain lesion segmentation. Medical Image Analysis
  36,  61--78 (2017)

\bibitem{kingma2014adam}
Kingma, D., Ba, J.: Adam: A method for stochastic optimization. arXiv preprint
  arXiv:1412.6980  (2014)

\bibitem{long2015fully}
Long, J., Shelhamer, E., Darrell, T.: Fully convolutional networks for semantic
  segmentation. In: Proceedings of the IEEE Conference on Computer Vision and
  Pattern Recognition. pp. 3431--3440 (2015)

\bibitem{menze2015multimodal}
Menze, B.H., Jakab, A., Bauer, S., Kalpathy-Cramer, J., Farahani, K., Kirby,
  J., Burren, Y., Porz, N., Slotboom, J., Wiest, R., et~al.: The multimodal
  brain tumor image segmentation benchmark (brats). IEEE transactions on
  medical imaging  34(10),  1993--2024 (2015)

\bibitem{poudel2016recurrent}
Poudel, R.P., Lamata, P., Montana, G.: Recurrent fully convolutional neural
  networks for multi-slice mri cardiac segmentation. arXiv preprint
  arXiv:1608.03974  (2016)

\bibitem{redmon2016you}
Redmon, J., Divvala, S., Girshick, R., Farhadi, A.: You only look once:
  Unified, real-time object detection. In: Proceedings of the IEEE Conference
  on Computer Vision and Pattern Recognition. pp. 779--788 (2016)

\bibitem{ronneberger2015u}
Ronneberger, O., Fischer, P., Brox, T.: U-net: Convolutional networks for
  biomedical image segmentation. In: International Conference on Medical Image
  Computing and Computer-Assisted Intervention. pp. 234--241. Springer (2015)

\bibitem{srivastava2014dropout}
Srivastava, N., Hinton, G.E., Krizhevsky, A., Sutskever, I., Salakhutdinov, R.:
  Dropout: a simple way to prevent neural networks from overfitting. Journal of
  Machine Learning Research  15(1),  1929--1958 (2014)

\end{thebibliography}
\end{document}